\documentclass[letterpaper, 10 pt, conference]{ieeeconf}  

\IEEEoverridecommandlockouts                              

\usepackage{cite}
\usepackage{amsmath,amssymb,amsfonts}
\usepackage{algpseudocode}
\usepackage{graphicx}
\usepackage{textcomp}
\usepackage{xcolor}
\usepackage{pifont}
\usepackage{caption}
\usepackage{hyperref}
\usepackage{multirow}

\title{\LARGE \bf
Adaptive Graduated Non-Convexity for Pose Graph Optimization
}

\author{Seungwon Choi, Wonseok Kang, Jiseong Chung, Jaehyun Kim, and Tae-wan Kim*
\thanks{* Corresponding Author}
\thanks{The authors are with the Department of Naval Architecture and Ocean Engineering, Seoul National University (SNU), Seoul 08826, Republic of Korea. \texttt{\{csw3575, eoid361, dntksdmfwmf, jaedalong, taewan\}@snu.ac.kr}
}
}

\begin{document}

\maketitle
\thispagestyle{empty}
\pagestyle{empty}

\begin{abstract}

We present a novel approach to robust pose graph optimization based on Graduated Non-Convexity (GNC). Unlike traditional GNC-based methods, the proposed approach employs an adaptive shape function using B-spline to optimize the shape of the robust kernel. This aims to reduce GNC iterations, boosting computational speed without compromising accuracy. When integrated with the open-source riSAM algorithm, the method demonstrates enhanced efficiency across diverse datasets. Accompanying open-source code aims to encourage further research in this area. \href{https://github.com/SNU-DLLAB/AGNC-PGO}{https://github.com/SNU-DLLAB/AGNC-PGO}

\end{abstract}

\section{Introduction}

Simultaneous Localization and Mapping (SLAM) is a fundamental research topic in robotics, as it contributes to providing essential information for the navigation of autonomous robotic systems. SLAM primarily consists of two modules: the front-end, responsible for processing sensor data, and the back-end, tasked with estimating the sensor trajectory by solving a non-linear least squares problem. Specifically, the back-end conducts pose graph optimization with loop closure factors based on the data association results from the front-end, to ensure global consistency in both trajectory and map representations. Given the inherent ambiguities in data association and potential measurement errors that can produce outliers, the performance of pose graph optimization hinges on how effectively it handles these outliers.

In this study, we introduce a robust pose graph optimization method based on Graduated Non-Convexity (GNC) \cite{black1996unification}. Unlike previous GNC-based approaches that rely on a predetermined graduation rule to adjust the shape of the robust kernel, we propose a novel adaptive shape function using B-spline. This function is designed to efficiently adjust the shape of the robust kernel, considering the relationship between the shape factor and the convexity of the kernel, with the aim of minimizing GNC iterations to potentially enhance computational speed without sacrificing accuracy. 

We applied our method to the GNC-based open-source pose graph optimization algorithm, robust incremental smoothing and mapping (riSAM) \cite{mcgann2023robust}. Through performance evaluations on various datasets, we demonstrate that our approach could enhance computational efficiency without compromising accuracy. Furthermore, we release the code from our experiments as open-source to support future research.

The remainder of this manuscript is structured as follows: Section II offers a brief overview of established robust pose graph optimization techniques. Section III describes the methodology introduced in this work. Section IV provides a comparative analysis with the existing method. Concluding remarks are presented in Section V.

\begin{figure}[!t]
    \includegraphics[width=0.486\textwidth]{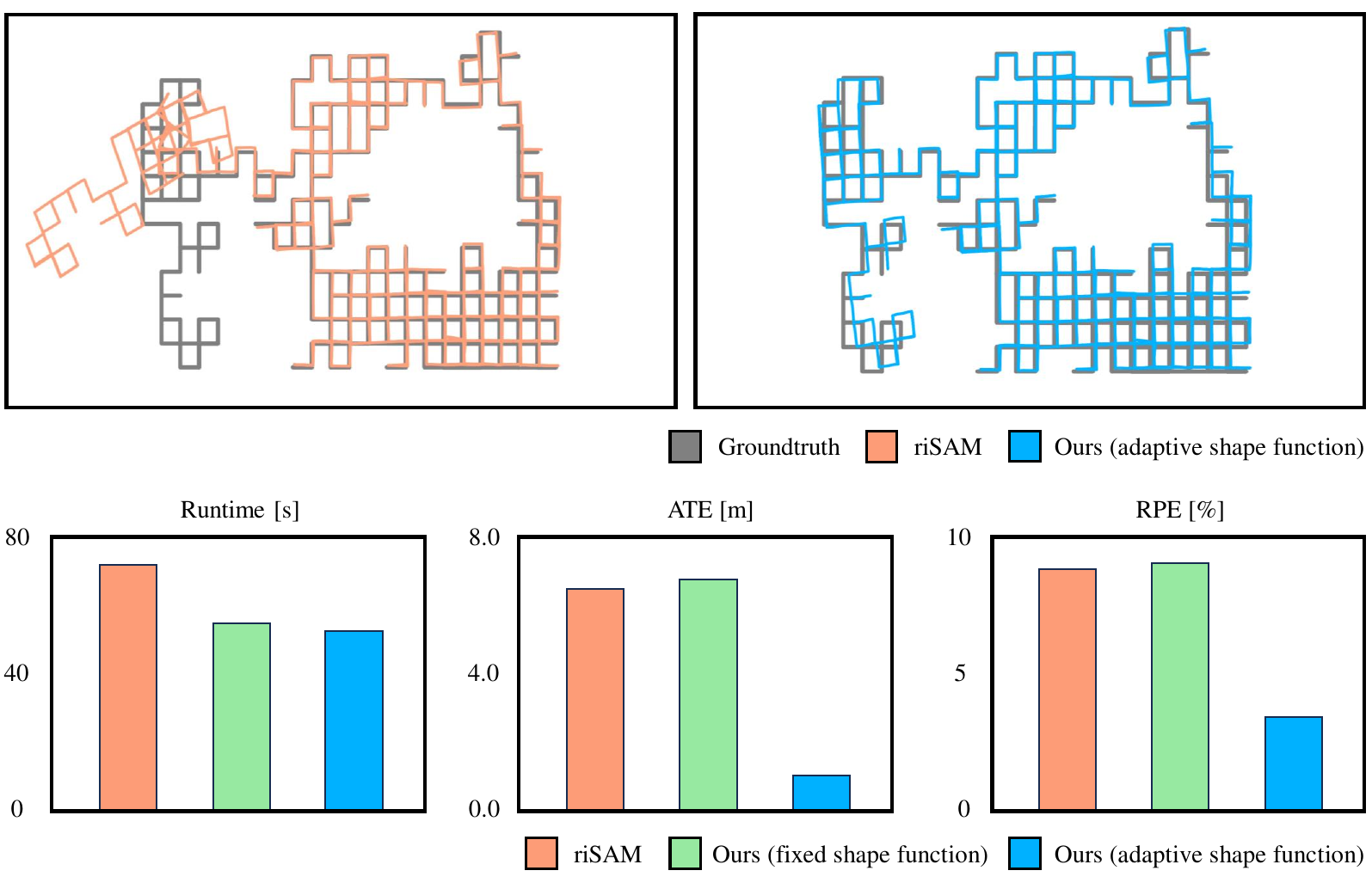}
    \caption{Results using the Manhattan3500 dataset with introduced noise and challenging outlier loops. The top images illustrate the ground truth trajectory, marked by the gray line, with the results of the riSAM algorithm on the left and the outcomes from our method (\textit{adaptive shape function}) on the right. The graph below compares the Absolute Trajectory Error (ATE), Relative Pose Error (RPE), and runtime derived from both the riSAM and our proposed methods.}
    \label{fig:first_page}
\end{figure}


\section{Related Work}

\begin{figure*}[t]
    \includegraphics[width=\textwidth]{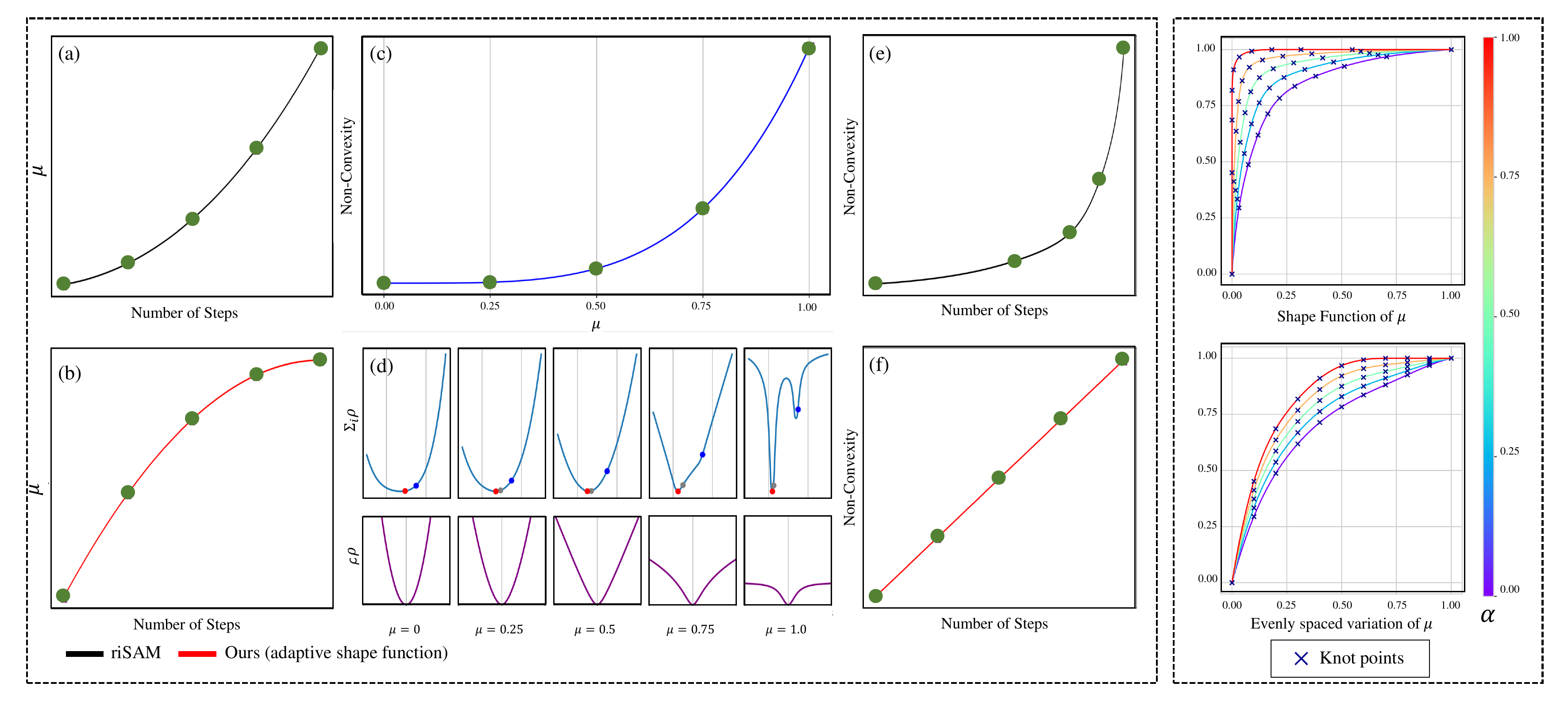}
    \caption{(a) Control parameter $\mu$ value for iteration steps in riSAM, which changes according to Eq. \eqref{eq:SIG}. The early increment is small, while the later increment is large. (b) Control parameter $\mu$ value for iteration steps in the proposed method. The early increment is large, while the later increment is small. (c) The non-convexity of the original cost function for the value of $\mu$. (d) Example of GNC with varying $\mu$ solving a linear regression problem with outliers using the GNC algorithm. From left to right, the cost (top) and the kernel function (bottom) obtained by applying $\mu={0,0.25,0.5,0.75,1.0}$ are depicted. The blue dot represents the initial guess, the gray dot represents the updated value from the previous stage, and the red dot represents the updated value from the current stage. (e) The non-convexity for iteration steps in riSAM. Since both the value of $\mu$ and non-convexity increases exponentially, the non-convexity for the iteration step increases even more exponentially. (f) The non-convexity for iteration steps in the proposed method. While the non-convexity increases exponentially, the shape function of $\mu$ is concave, so the non-convexity for the iteration step increases in a linear or near-linear fashion. (g) The shape function of $\mu$ is defined by B-spline. As the Mahalanobis distance varies between $\boldsymbol{\chi}^2(0.25)$ and $\boldsymbol{\chi}^2(0.9)$, the shape function changes according to the interpolation parameter $\alpha$. (h) Evenly spaced variation of $\mu$. It implies iteration steps on the horizontal axis, and the value of $\mu$ for iteration steps increases in a concave manner.}
    \label{fig:overview}
\end{figure*}

Pose graph optimization is a non-linear least squares problem, and the accuracy of its solutions depends on the careful management of outliers. In this context, various methodologies have been proposed to address these challenges.

Building on the principle of data consensus within the graph, Latif \textit{et al.}\cite{latif2012realizing} introduce the iRRR algorithm, emphasizing clustering and consistency-checking to reject false loop closures. Similarly, Sünderhauf \textit{et al.}\cite{sunderhauf2012switchable} propose an approach using switchable constraints, offering a dynamic mechanism to evaluate and potentially dismiss certain loop closures, underscoring the significance of data consensus in the graph. 

Other studies introduce methods based on mixture models for multi-modal cases. Olson \textit{et al.}\cite{olson2013inference} present an alternative to the traditional sum-mixture approach, selecting the max values from two probability density functions to achieve a similar distribution with reduced computational overhead. Pfeifer \textit{et al.}\cite{pfeifer2021advancing} offer a hybrid formulation combining both sum and max mixtures (max-sum-mixture), which, despite requiring more computation, incorporates non-linear terms and demonstrates enhanced accuracy.


In the realm of robust kernel-based techniques, methods \cite{huber2004robust} based on the M-estimator\cite{zhang1997parameter} are explored. Among these, Chebrolu \textit{et al.}\cite{chebrolu2021adaptive} introduce robust kernels with a general and adaptive loss function\cite{Barron} for non-linear least squares problems, emphasizing automatic kernel tuning based on residual distribution. Their approach broadens existing robust kernels and delivers enhanced accuracy without manual parameter adjustments.

From another perspective, Yang \textit{et al.}\cite{yang2020graduated}. introduce graduated non-convexity (GNC)\cite{GNC} to address outlier vulnerabilities in semidefinite programming (SDP) and sums-of-squares (SOS) relaxations, leveraging the Black-Rangarajan duality to outperform techniques like RANSAC\cite{fischler1981random}. Building on this, McGann \textit{et al.}\cite{mcgann2023robust} extend the GNC approach to online incremental SLAM, presenting the robust incremental Smoothing and Mapping (riSAM) algorithm. Their method maintains online efficiency and outperforms existing SLAM techniques.

In this paper, we propose a method based on GNC. Distinctively from existing methods, we introduce a graduation approach that considers the relationship between the control parameter of the robust kernel and non-convexity. Furthermore, instead of changing the kernel based on predetermined steps, we suggest an adaptive method that varies the kernel according to statistical significance.

\begin{figure*}[!ht]
    \includegraphics[width=\textwidth]{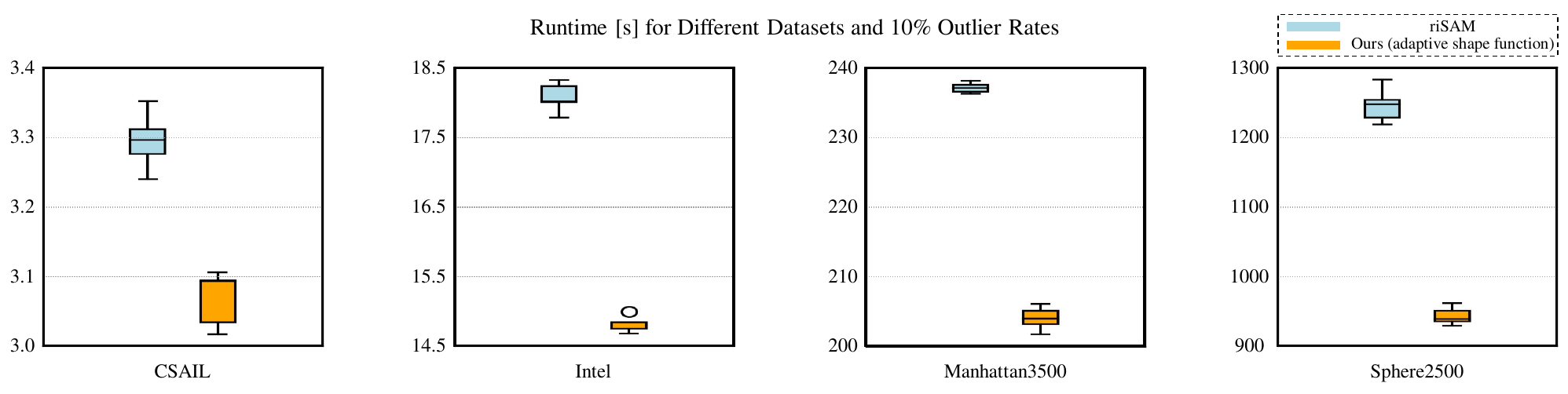}
    \caption{Runtime comparison between our algorithm and riSAM over four datasets. The graphs represent the runtime results when each dataset is subjected to 10\% outliers and run five times.
    \textit{Note: The y-axis scale varies for each graph in the top section}.}
    \label{fig:experimental_result}
\end{figure*}

\section{Methodology}

In this section, we provide a brief overview of GNC and subsequently describe our approach to enhancing efficiency within GNC.

\subsection{Graduated Non-Convexity (GNC)}

In optimization, GNC starts with a simple convex problem and gradually increases the non-convexity to recover the original problem. It eliminates the need for an initial guess and increases the likelihood of converging to the global minimum. 
Traditional robust kernels, such as GM (Geman-McClure), faced challenges due to their lack of convexity characteristics across the entire domain. In contrast, McGann \textit{et al.} \cite{mcgann2023robust} proposed the Scale Invariant Graduated (SIG) kernel, which maintains convexity properties throughout its domain. Building on this, our work also employs the SIG kernel to ensure stable optimization.

The SIG kernel \cite{mcgann2023robust} is defined as:
\begin{equation}
    \label{eq:SIG}
    \rho (r;\mu)= \frac{1}{2} \frac{c^2r^2}{c^2 + (r^2)^\mu}, \quad 0 \le \mu \le 1 
\end{equation}
The non-convexity is controlled by a control parameter, $\mu$. The kernel is quadratic when $\mu=0$, and becomes Geman-McClure when $\mu=1$. In previous works \cite{yang2020graduated}, \cite{mcgann2023robust}, the rise in non-convexity begins with a slight increase and then gradually becomes higher.
However, as shown in Fig. \ref{fig:overview}, it is observed that a non-convexity of a function doesn't always linearly correlate with $\mu$. Therefore, this graduation method presented in previous studies increases the non-convexity of the function exponentially. 

To make GNC operate efficiently, we propose a graduation approach that can linearly increase non-convexity. Our method initially increases the value of $\mu$ substantially and then gradually reduces the rate of increase.

\subsection{Adaptive Shape Function}
The \textit{shape function} determines the $\mu$ value for each GNC graduation step, and this $\mu$ value, in turn, determines the non-convexity of the robust kernel. To mitigate the influence of outliers, we propose a method that adaptively adjusts the shape function based on the statistical significance of each factor. The aim is to assign a larger initial $\mu$ value to factors deemed as outliers, thereby enhancing the robustness of the optimization. In order to achieve flexible adjustment of the \textit{shape function} with a minimal number of parameters, we derive the adaptive shape function using B-spline.

The B-spline curve is defined as follows:
\begin{equation}
    \mathbf{s}(u)=\sum_{i=0}^{n}{N_i^k(u)\mathbf{d}_i}    
\end{equation}
where $\boldsymbol{\mathrm{d}}=(\mathbf{d}_i)_{i=0}^{n-1}$, $\mathbf{d}_i\in \mathbb{R}^m$ are the control points which are real-valued points existing in an $m$-dimensional space and $N_i^k$ are B-spline basis where the degree of B-spline curve is $k$.

    The B-spline basis functions are defined using a recursive formulation \cite{de1978practical}. When $k=0$, the normalized B-spline basis is defined as:
    \begin{equation}
        N_i^0(u)=
        \begin{cases}
        1 &  u_i \leq u < u_{i+1} \\
        0 & \text{otherwise}
        \end{cases}
    \end{equation}
    Also, when $k>0$:
    \begin{multline}
        \quad\quad\quad N_i^k(u) = \frac{u - u_i}{u_{i+k} - u_i} N_i^{k-1}(u) \\
        + \frac{u_{i+k+1} - u}{u_{i+k+1} - u_{i+1}} N_{i+1}^{k-1}(u) \quad\quad\quad
    \end{multline}
    where the knot vector \(\boldsymbol{\mathrm{u}} = \{u_0, u_1, \ldots, u_{n+k+1}\}\) \cite{PieglTiller1997}. 
    The $u \in [0,1]$ represents a specific point on the curve and can also be considered a time point, in the context of curve parameterization. For example, $\mathbf{s}(u=0.25)$ represents the point on the curve corresponding to $u=0.25$ within a spline whose parameter space is defined over the interval $[0,1]$.
    
    We construct the shape function such that its initial gradient is large, and it decreases as it progresses. Additionally, for an adaptive configuration, we design functions according to various initial gradients. The shape function can be adaptively controlled by an \textit{interpolation parameter} $\alpha$, shown as Fig. \ref{fig:overview}, between $\alpha=0$ and  $\alpha=1$.
    
    We classify factors as inliers if the Mahalanobis distance is less than the point where the cumulative distribution function of the chi-square distribution reaches 0.25. Conversely, if the Mahalanobis distance exceeds the point where this distribution reaches 0.9, we categorize them as outliers. 
    
    The $\alpha$ is defined as:
    \begin{equation}
    \alpha = 
        \begin{cases} 
            0 & \text{if } m < \boldsymbol{\chi}^2(0.25) \\
            \frac{m - \boldsymbol{\chi}^2(0.25)}{\boldsymbol{\chi}^2(0.9) - \boldsymbol{\chi}^2(0.25)}  & \text{if } \boldsymbol{\chi}^2(0.25) \le m < \boldsymbol{\chi}^2(0.9) \\
            1 & \text{if }  m \ge \boldsymbol{\chi}^2(0.9)
        \end{cases}
        \label{eq: alpha}
    \end{equation},
    where $m$ is the Mahalanobis distance of a factor. For inliers, we apply $\alpha=0$ to ensure a gradual increase in $\mu$. For outliers, $\alpha=1$ is applied to facilitate a rapid increase in $\mu$. For the rest, an interpolated shape function is applied based on the value of $\alpha$.

\subsection{GNC with Adaptive Shape Function}

    Leveraging the adaptive shape function, we propose an effective method for selecting $\mu$ during the optimization process of GNC. Initially, we compute the Mahalanobis distance for each factor and categorize them based on their distance into inliers, outliers, or other classifications. Subsequently, for each case, we calculate $\alpha$ according to Eq.\eqref{eq: alpha} and construct a shape function. The value of $\mu$ is then increased by this function.
    We set the kernel function Eq.\eqref{eq:SIG} with $c=3$ as we assume that inliers are within 3$\sigma$.
    Lastly, after optimization, we recompute the Mahalanobis distance to finally classify whether it's an inlier or outlier. Based on this classification, we adjust the initial value of $\mu$, denoted as $\mu_{\text{init}}$, for the next optimization iteration.

\section{Experiments}

In this section, we present the evaluation results of our proposed algorithm on pose graph optimization datasets. In our first experiment, we utilize the \textit{CSAIL} \cite{csaildata}, \textit{Intel} \cite{intelmanhattandataset}, \textit{Manhattan3500}\cite{manhattandataset}, \cite{intelmanhattandataset}, and \textit{Sphere2500} \cite{spheredataset} datasets, which were also employed in riSAM, to demonstrate that our algorithm can maintain data association accuracy while enhancing computational efficiency. In our second experiment, by generating a more challenging dataset derived from the ground truth of the Manhattan3500 dataset, we illustrate the incremental improvements in SLAM results attributed to our proposed graduation approach and adaptive shape function.

\begin{table}[]
\caption{Runtime comparison between our algorithm and riSAM using \textit{CSAIL, Intel, Manhattan3500, Sphere2500} dataset for each outlier ratio. The last column implies the runtime ratio of the proposed algorithm to the riSAM algorithm.}
\label{table:results1}
\begin{tabular}{c|c|cc|c}
\hline
\multicolumn{2}{c|}{Runtime (s)}                                                             & \multirow{2}{*}{riSAM} & \multirow{2}{*}{Ours}         & \multirow{2}{*}{\begin{tabular}[c]{@{}c@{}}Runtime ratio\\ (Ours / riSAM)\end{tabular}} \\ \cline{1-2}
Dataset                    & \begin{tabular}[c]{@{}c@{}}Outlier\\ Ratio (\%)\end{tabular} &                        &                               &                                                                                      \\ \hline
\multirow{3}{*}{CSAIL}     & 10                                                           & 3.3528                 & \multicolumn{1}{c|}{3.0374}   & 0.906                                                                                \\
                           & 30                                                           & 4.7476                 & \multicolumn{1}{c|}{3.9662}   & 0.835                                                                                \\
                           & 50                                                           & 8.291                  & \multicolumn{1}{c|}{6.2908}   & 0.759                                                                                \\ \hline
\multirow{3}{*}{Intel}     & 10                                                           & 17.6464                & \multicolumn{1}{c|}{14.7472}  & 0.836                                                                                \\
                           & 30                                                           & 45.1486                & \multicolumn{1}{c|}{35.829}   & 0.794                                                                                \\
                           & 50                                                           & 140.021                & \multicolumn{1}{c|}{105.792}  & 0.756                                                                                \\ \hline
\multirow{3}{*}{Manhattan} & 10                                                           & 232.9804               & \multicolumn{1}{c|}{198.4186} & 0.852                                                                                \\
                           & 30                                                           & 893.6256               & \multicolumn{1}{c|}{741.713}  & 0.830                                                                                \\
                           & 50                                                           & 3732.312               & \multicolumn{1}{c|}{3060.65}  & 0.820                                                                                \\ \hline
\multirow{3}{*}{Sphere}    & 10                                                           & 1216.06                & \multicolumn{1}{c|}{1031.38}  & 0.848                                                                                \\
                           & 30                                                           & 7061.93                & \multicolumn{1}{c|}{5839.47}  & 0.827                                                                                \\
                           & 50                                                           & 33948.8                & \multicolumn{1}{c|}{28072.7}  & 0.827                                                                                \\ \hline
\end{tabular}
\end{table}

In the first experiment, we use four distinct datasets and artificially introduce outliers into the loop closure factors. These outliers represent cases where distant poses are mistakenly identified as the same pose, leading to erroneous data associations. We evaluate our proposed algorithm and the riSAM method on these modified datasets, focusing on metrics such as total runtime, recall, and precision. As presented in Table \ref{table:results1}, our method consistently offers improved runtime efficiency across all datasets and outlier ratios. This trend is visually emphasized in Fig. \ref{fig:experimental_result}, which provides a clear depiction of our algorithm's superior computational performance. Both algorithms yield identical values for recall and precision across most datasets. However, as presented in Table \ref{table:results_extended}, for the \textit{Sphere2500} dataset, our method displays a slightly better precision and recall.

\begin{table}
\centering
\caption{Precision and recall comparison between our algorithm and riSAM using \textit{Sphere2500} dataset, incorporating outliers at rates of 10\%, 30\%, and 50\%.}
\label{table:results_extended}
\begin{tabular}{c|ccc|ccc}
\hline\hline
Metric & \multicolumn{3}{c|}{Precision} & \multicolumn{3}{c}{Recall}  \\
\hline
Outlier rate (\%) & 10 & 30 & 50 & 10 & 30 & 50 \\
\hline
riSAM  & 1.0 & 1.0 & 0.999 & 0.998 & 0.997 & 0.985 \\
\textbf{Ours}   & \textbf{1.0} & \textbf{1.0} & \textbf{1.0} & \textbf{0.999} & \textbf{0.998} & \textbf{0.998} \\
\hline
\end{tabular}
\end{table}

In the second experiment, we evaluate the performance of the algorithms in the presence of noise in loop closure measurements. For this purpose, we generate odometry edges and loop closure edges from the ground truth poses of the \textit{Manhattan3500} dataset. We then introduce a relatively large amount of noise to some of the loop closure measurements. We compare the performance of riSAM and our algorithm on this artificially corrupted data. Specifically, to ascertain the impact of the proposed graduation approach and adaptive shape function on performance enhancement, we examine the performance in two modes: \textit{mode 1}, which only applies the graduation approach with a fixed $\alpha$ value of 0.5, and \textit{mode 2}, which employs both the graduation approach and the adaptive shape function. As shown in the results in Fig. \ref{fig:first_page}, the graduation approach proposed in this paper enables achieving trajectory estimation accuracy comparable to riSAM with fewer steps, thereby decreasing total runtime. Furthermore, the introduction of the adaptive shape function demonstrates increased robustness against outliers.

\section{Discussion and Future Work}

The results presented in this study highlight the potential of the proposed algorithm to improve the computational efficiency and robustness of pose graph optimization. The method consistently outperformed riSAM, especially in scenarios with a high percentage of outliers, which emphasizes the effectiveness of the graduated approach and the adaptive shape function. In particular, the advantage of the adaptive shape function in enhancing robustness to outliers is shown to be an important contribution to the SLAM domain.

While the experiments provide promising results on both artificially corrupted and real-world datasets, there is still room for further exploration. This study mainly focuses on a few datasets, and the robustness and efficiency of the method in various real-world scenarios need more research, which will be done in future studies.

\newpage

{\small
\bibliographystyle{IEEEtran}
\bibliography{egbib}

\begin{thebibliography}{10}
\providecommand{\url}[1]{#1}
\csname url@rmstyle\endcsname
\providecommand{\newblock}{\relax}
\providecommand{\bibinfo}[2]{#2}
\providecommand\BIBentrySTDinterwordspacing{\spaceskip=0pt\relax}
\providecommand\BIBentryALTinterwordstretchfactor{4}
\providecommand\BIBentryALTinterwordspacing{\spaceskip=\fontdimen2\font plus
\BIBentryALTinterwordstretchfactor\fontdimen3\font minus \fontdimen4\font\relax}
\providecommand\BIBforeignlanguage[2]{{%
\expandafter\ifx\csname l@#1\endcsname\relax
\typeout{** WARNING: IEEEtran.bst: No hyphenation pattern has been}%
\typeout{** loaded for the language `#1'. Using the pattern for}%
\typeout{** the default language instead.}%
\else
\language=\csname l@#1\endcsname
\fi
#2}}

\bibitem{black1996unification}
M.~J. Black and A.~Rangarajan, ``On the unification of line processes, outlier rejection, and robust statistics with applications in early vision,'' \emph{International journal of computer vision}, vol.~19, no.~1, pp. 57--91, 1996.

\bibitem{mcgann2023robust}
D.~McGann, J.~G. Rogers, and M.~Kaess, ``Robust incremental smoothing and mapping (risam),'' in \emph{2023 IEEE International Conference on Robotics and Automation (ICRA)}.\hskip 1em plus 0.5em minus 0.4em\relax IEEE, 2023, pp. 4157--4163.

\bibitem{latif2012realizing}
Y.~Latif, C.~Cadena, and J.~Neira, ``Realizing, reversing, recovering: Incremental robust loop closing over time using the irrr algorithm,'' in \emph{2012 IEEE/RSJ International Conference on Intelligent Robots and Systems}.\hskip 1em plus 0.5em minus 0.4em\relax IEEE, 2012, pp. 4211--4217.

\bibitem{sunderhauf2012switchable}
N.~S{\"u}nderhauf and P.~Protzel, ``Switchable constraints for robust pose graph slam,'' in \emph{2012 IEEE/RSJ International Conference on Intelligent Robots and Systems}.\hskip 1em plus 0.5em minus 0.4em\relax IEEE, 2012, pp. 1879--1884.

\bibitem{olson2013inference}
E.~Olson and P.~Agarwal, ``Inference on networks of mixtures for robust robot mapping,'' \emph{The International Journal of Robotics Research}, vol.~32, no.~7, pp. 826--840, 2013.

\bibitem{pfeifer2021advancing}
T.~Pfeifer, S.~Lange, and P.~Protzel, ``Advancing mixture models for least squares optimization,'' \emph{IEEE Robotics and Automation Letters}, vol.~6, no.~2, pp. 3941--3948, 2021.

\bibitem{huber2004robust}
P.~J. Huber, \emph{Robust statistics}.\hskip 1em plus 0.5em minus 0.4em\relax John Wiley \& Sons, 2004, vol. 523.

\bibitem{zhang1997parameter}
Z.~Zhang, ``Parameter estimation techniques: A tutorial with application to conic fitting,'' \emph{Image and vision Computing}, vol.~15, no.~1, pp. 59--76, 1997.

\bibitem{chebrolu2021adaptive}
N.~Chebrolu, T.~L{\"a}be, O.~Vysotska, J.~Behley, and C.~Stachniss, ``Adaptive robust kernels for non-linear least squares problems,'' \emph{IEEE Robotics and Automation Letters}, vol.~6, no.~2, pp. 2240--2247, 2021.

\bibitem{Barron}
J.~Barron, ``A general and adaptive robust loss function,'' 06 2019, pp. 4326--4334.

\bibitem{yang2020graduated}
H.~Yang, P.~Antonante, V.~Tzoumas, and L.~Carlone, ``Graduated non-convexity for robust spatial perception: From non-minimal solvers to global outlier rejection,'' \emph{IEEE Robotics and Automation Letters}, vol.~5, no.~2, pp. 1127--1134, 2020.

\bibitem{GNC}
A.~Blake and A.~Zisserman, \emph{Visual Reconstruction}.\hskip 1em plus 0.5em minus 0.4em\relax Cambridge, MA, USA: MIT Press, 1987.

\bibitem{fischler1981random}
M.~A. Fischler and R.~C. Bolles, ``Random sample consensus: a paradigm for model fitting with applications to image analysis and automated cartography,'' \emph{Communications of the ACM}, vol.~24, no.~6, pp. 381--395, 1981.

\bibitem{de1978practical}
C.~De~Boor and C.~De~Boor, \emph{A practical guide to splines}.\hskip 1em plus 0.5em minus 0.4em\relax springer-verlag New York, 1978, vol.~27.

\bibitem{PieglTiller1997}
L.~Piegl and W.~Tiller, \emph{The NURBS Book}, 2nd~ed.\hskip 1em plus 0.5em minus 0.4em\relax Berlin: Springer-Verlag, 1997.

\bibitem{csaildata}
L.~Carlone, R.~Aragues, J.~A. Castellanos, and B.~Bona, ``A fast and accurate approximation for planar pose graph optimization,'' \emph{The International Journal of Robotics Research}, vol.~33, no.~7, pp. 965--987, 2014.

\bibitem{intelmanhattandataset}
L.~Carlone and A.~Censi, ``From angular manifolds to the integer lattice: Guaranteed orientation estimation with application to pose graph optimization,'' \emph{IEEE Transactions on Robotics}, vol.~30, no.~2, pp. 475--492, 2014.

\bibitem{manhattandataset}
E.~Olson, J.~Leonard, and S.~Teller, ``Fast iterative alignment of pose graphs with poor initial estimates,'' in \emph{Proceedings 2006 IEEE International Conference on Robotics and Automation, 2006. ICRA 2006.}\hskip 1em plus 0.5em minus 0.4em\relax IEEE, 2006, pp. 2262--2269.

\bibitem{spheredataset}
L.~Carlone, R.~Tron, K.~Daniilidis, and F.~Dellaert, ``Initialization techniques for 3d slam: A survey on rotation estimation and its use in pose graph optimization,'' in \emph{2015 IEEE international conference on robotics and automation (ICRA)}.\hskip 1em plus 0.5em minus 0.4em\relax IEEE, 2015, pp. 4597--4604.

\end{thebibliography}
}

\end{document}